\documentclass{elsarticle}

\usepackage{hyperref}

\usepackage{multirow}
\journal{Journal of \LaTeX\ Templates}

\begin{document}

\begin{frontmatter} 

\title{Optimized clothes segmentation to boost gender classification in unconstrained scenarios}

\author{D. Freire-Obreg\'on\footnote{Corresponding author e-mail address: david.freire@ulpgc.es}, M. Castrill\'on-Santana and J. Lorenzo-Navarro}

\address{SIANI - Universidad de Las Palmas de Gran Canaria (ULPGC)\\Las Palmas - Spain}







\begin{abstract}
Several applications require demographic information of ordinary people in unconstrained scenarios. This is not a trivial task due to significant human appearance variations.
In this work, we introduce trixels for clustering image regions, enumerating their advantages compared to superpixels. The classical GrabCut algorithm is later modified to segment trixels instead of pixels in an unsupervised context.
Combining with face detection lead us to a clothes segmentation approach close to real time. 
The study uses 
the challenging Pascal VOC dataset for  segmentation evaluation experiments. A final experiment analyzes the fusion of clothes features with state-of-the-art gender classifiers in ClothesDB, revealing a significant performance improvement in gender classification.  
\end{abstract}


\end{frontmatter}


\section{Introduction}
\label{sec:Introduction}

The analysis of humans has been a recurrent research topic in the history of humankind. Regarding it, several science-related fields have addressed the structure, the evolution or other general features of human population. One relevant aspect which describes humans is their appearance. 

Although there are attributes of human appearance that are chosen by people, others can not be consciously defined. Physical traits (e.g., hair color or beard style) and adhered human characteristics (e.g., the kind of clothes or the presence of tattoos) are elements that can be defined by humans. Among the non-definable ones, we can mention height, ethnicity, age, and so on. Dantcheva et al. \cite{Dantcheva16} claimed that these attributes can be used in a fusion framework to improve a primary biometric system, i.e., fusing clothes with gender information.



In this paper, our final goal is to automatically estimate gender. Gender classification (GC) is a two class learning problem; male (M) / female (F). Although it may seem a simple problem to handle, there are multiple factors that engage this classification task. As can be seen in section \ref{sec:stateart}, most of the techniques used in computer vision for GC are based on facial features. However, GC from face images is not
a valid approach in many situations, especially under uncontrolled environments such as happens in video surveillance data, where resolution, pose, lighting or occlusive elements (glasses and hair) can hinder 
correct gender identification.
Due to these difficulties, as happens for face recognition \cite{Bhardwaj15-icb}, humans often explore alternative cues for GC through contextual elements such as 
clothing or hair \cite{Li12}.
Despite these evidences, to integrate contextual and facial information is not a trivial task. This may be because of the difficulties that are present in automatic clothes segmentation: cluttered backgrounds, complex human pose, or clothes properties. Hence, we claim that there is a need for the development of robust techniques to extract contextual elements that can be combined with the traditional facial cues.


\begin{figure}[tbp]
  \centering
  \includegraphics[width=100mm]{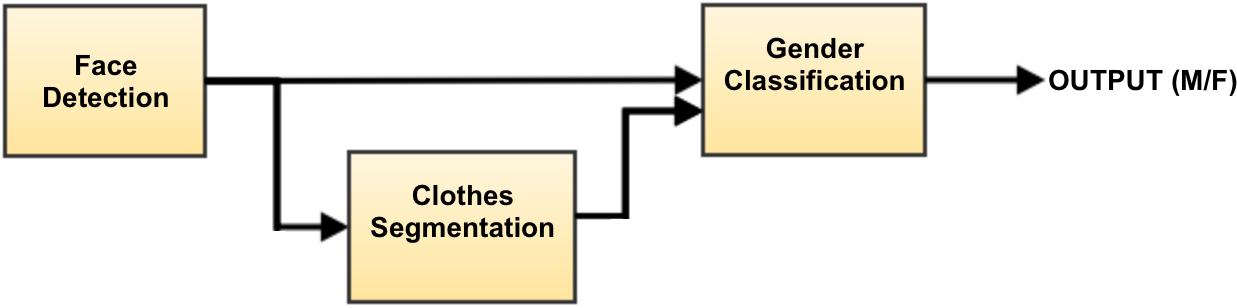}
  \caption{General proposal overview. Face detection automates the segmentation process. Then, facial and clothes features are fused for GC.}
  \label{fig:genderflow}
\end{figure}

Body can provide interesting cues about the gender of a person; accessories or type of clothes are relevant in the decision making process. However,
there are two challenging cultural aspects to deal with; on the one hand, people of the same gender may choose different style of clothes
depending on their personality or the event they are attending. On the other hand, clothing styles worn by both males and females may be 
somewhat similar \cite{Choon12}. 

Our hypothesis is that we can boost GC accuracy by using soft biometric information. For this task, clothes information must be extracted before facing GC. In the last decade, the GrabCut \cite{Rother04} technique has been widely used in applications for mostly interactive foreground extraction \cite{Haasch05, Vaiapury10}. Unfortunately, this algorithm is time-consuming for real time applications due to multiple factors; the algorithm converge slowly because of a complex scenario or large images. Besides, human supervision is required to fit the area of interest. Because of this, we propose the use of an optimized segmentation approach based on the classical GrabCut technique. In this sense, Figure \ref{fig:genderflow} shows a general overview of our GC proposal. First, face detection is used to automatically guide the process of clothes segmentation. Then, clothes information and facial cues are both combined in order to classify the person gender (male/female).


This paper extends \cite{Freire14}, where this concept was initially presented. However, a new alternative is included to automatically setup the GrabCut trimap and an updated state-of-the-art facial based gender classifier is integrated, showing an evident GC improvement in terms of accuracy.
Our major contributions are as follows: 1) the use of trixels in order to simplify the image data into perceptually meaningful atomic regions, 2) the adaptation of the classical Grabcut to segment trixels instead of pixels reducing significantly the processing cost, 3) an improvement of clothes segmentation quality compared to our previous results, and, 4) the successful combination of an state-of-the-art gender classifier with clothing features, boosting GC.


The paper is organized in seven sections. Section \ref{sec:stateart} looks at the related work. In section \ref{sec:trixels}, trixels are presented as well as their relation with superpixels. The overall segmentation process is described in section \ref{sec:segmentation}. In section \ref{sec:setup} the feature extraction for GC is described. Then, segmentation and GC experiments are reported in section \ref{sec:experimental1}. Finally, conclusions are drawn in section \ref{sec:conclusions}.

\section{Related work}
\label{sec:stateart}
GC is a trending challenge in the field of computer vision. Related work makes use of techniques that can be grouped into geometric or appearance based approaches. Furthermore, facial features can be analyzed considering the whole face \cite{Rai14} or just focusing on specific facial regions \cite{Wu11} in order to determine the gender of a person with greater precision.
Recently, Rai and Khanna \cite{Rai14} combined a support vector machine (SVM) and Gabor filters for GC and they achieved an accuracy of 98.4\% on the FERET dataset. Although Wu and Lian \cite{Wu11} worked with normalized images too 
(CAS-PEAL database), they only considered facial components instead of the whole face (97\% accuracy). Moreover, these authors claimed that facial components provide a higher 
robustness against variations caused by facial alignment, illumination and occlusions. 
Despite these works, facial cues are combined with context information surrounding the subject in the design of new proposals in the field. 
Therefore, a pretty remarkable study about which human parts are relevant for GC
is presented by Li et al. \cite{Li12}. Their approach generates seven different classifiers (hair, forehead, eyes, nose, mouth, chin and the
upper chest part respectively).
For the FERET dataset, the best results combining the seven classifiers reports an accuracy of 93.4\%.
A more challenging proposal is presented by Shan \cite{Shan12} using
Local Binary Patterns (LBP) and SVM for GC on wild images. Shan proposal obtained an accuracy of 94.81\% on the LFW database. Also Castrill\'on et al. \cite{Castrillon16_arxiv} combined local descriptors computed on the facial area and its local context, and convolutional neural networks (CNN) for GC. They achieved an accuracy of 98.1\% on The Images of Groups dataset. 

Besides, several works have been carried out on body-based GC. 
Thus, Cao et al. \cite{Cao08} achieved an accuracy of 75\% considering frontal images from the MIT-CBCL database. An approach based on histograms of oriented gradients (HOG) was presented by Collins et al. 
\cite{Collins09} improving Cao's previous accuracy up to a 80.62\%. Nonetheless, the 82.4\% accuracy with unconstrained images of people
obtained by Bourdev et al. \cite{Bourdev11} outperforms previous authors. However, their evaluated dataset is reduced and not balanced.


Body segmentation is not a trivial task; the human pose, the illumination, or the oclusions hamper the segmentation process. Below, clothes features are combined with local context and facial information to boost GC. 

\section{Triangular superpixels}
\label{sec:trixels}

\begin{figure}[t]

\begin{minipage}[b]{1.0\linewidth}
  \centering
  \centerline{\includegraphics[width=12.0cm]{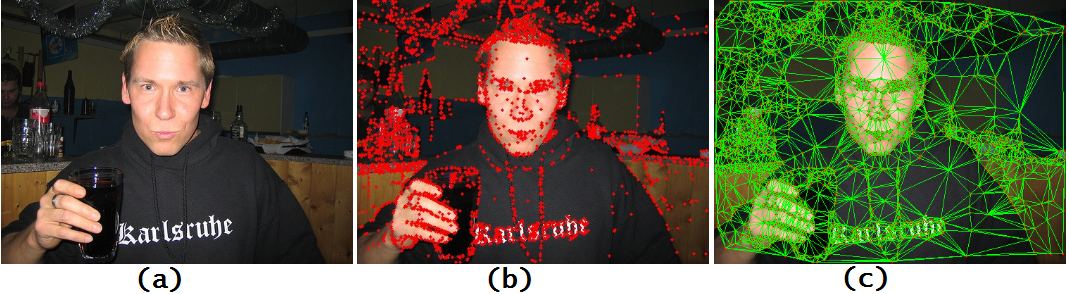}}
\end{minipage}
\caption{Process from the source image (a) to TriToM (c) through DMUM (b) \cite{Freire14}.}
\label{fig:tritomcreation}
\end{figure}

In the last decade, superpixels \cite{Ren03} have proved to be increasingly useful for applications such as image segmentation \cite{Achanta12}, sketonization \cite{Levinshtein13} and object localization \cite{Fulkerson09}. Superpixels adapt themselves to the image structure reducing its complexity and simplifying the management of the image content.
The basic fundamentals of trixels\footnote{https://github.com/davidfreire/Tritom} (triangular 
superpixels) are presented in this section. A trixel is defined as a computable atomic unit having a 
triangular shape. The trixel creation is a two-step process: 
\begin{itemize}
\item Firstly, a set of vertices are placed all over the image. Their placement is based on the classical definition of the Distance Transform (DT) where each point receives a value representing a point distance from the nearest object \cite{Rosenfeld68}. However, Ant\'on-Canal\'is et al. have redefined the term object \cite{Anton12} in their Distance from Unthresholded Magnitudes Maps (DMUM) algorithm, where the expression on the "distance to the nearest object" was replaced by the "distance to the nearest most relevant 
magnitude area" in a mapping function directly based on unthresholded magnitudes. Thus, DMUM provides a  map of the local minima located along the processed image and defines the trixel vertices (see Figure \ref{fig:tritomcreation}-b). 
\item Secondly, these vertices are the input of a Delaunay triangulation. Thus, pixels of the original image are 
grouped into a set of trixels which exhibit a homogeneity regarding the information contained in each subregion.
Once the Delaunay triangulation is computed, a manipulable trixel's mesh known as Trixel Topological Mesh (TriToM) composed by a finite number of trixels is generated (see Figure \ref{fig:tritomcreation}-c).
\end{itemize}

Although trixels exhibit a certain similarity with superpixels, it is a novel approach
which offers a dynamic and adaptable proposal to real-time applications \cite{Freire14}. As happens to superpixels, the trixels divide the image into a finite set of independent segments. 
It should be noticed that trixels fulfill some important superpixels properties. Achanta et al. \cite{Achanta12} described these general and desirable properties for each approach based on the idea of grouping of pixels: 
\begin{itemize}
\item The new atomic unit must conform precisely to the existing edges in the image. 
\item When they are generated in order to reduce the computational complexity of the image, should be fast calculation, memory-efficient and easy to manipulate.
\end{itemize}

Similarly to superpixels, a segmentation technique based on a directed graph can be proposed. In fact, our proposal considers each trixel as a node in a graph. 
Thus, the set of the created trixels form a mesh all over the image. On the other hand, unlike what happens with superpixels, trixels are not randomly placed in its first phase, neither the image is divided evenly (considering blocks of $n \times m$ pixels) without taking into account the image information during this first phase (as happens in superpixels \cite{Achanta12}). As mentioned above, the creation of the trixel mesh is based on the information provided by the processed image through DMUM. Contrary to superpixels, trixels present three neighbors at most, which facilitates the TriToM management. 
In this sense, the trixel mesh topology is based on the technique of the 
Delaunay triangulation and so, it inherits some interesting properties of this triangulation such as: 
1) the configuration stability that can be used to index the trixels formed through this technique is ensured, 2) inserting a new point into a Delaunay triangulation only affects those triangles whose circumscribed circles affect that point and, 3) the structure ensures the stability of 
the neighborhood of the triangles. The resulting triangulation exhibits a finer resolution, smaller trixels, in areas with details as contours and coarser resolution, bigger trixels, in homogeneous areas.



\section{TriToM-based clothes segmentation}
\label{sec:segmentation}

In this section, we address the image segmentation problem through the use of a new technique that facilitates the extraction of the object of interest regardless of the 
characteristics of the background where this object is placed. Indeed, our proposal may be adopted for different purposes. The keypoint is the GrabCut trimap definition. In our case, being interested in GC, we will opportunistically make use of the facial detection information to define the necessary trimaps to automatically guide both skin and clothes segmentation. By following this approach, there is no need for the segmentation process supervision by any human.

In order to speed-up GrabCut based segmentation \cite{Rother04}, the algorithm is adapted to process trixels instead of pixels. 
The aim is to reduce the algorithm processing cost maintaining a reasonable segmentation quality.
Using TriToM as input, 
the goal is to assign each trixel to a cluster that shares the same properties.

Contrary to what happens with superpixels, the segmentation process is not performed directly on the TriToM. In other words, each trixel is not processed at atomic level due to the following reasons:
\begin{itemize}
\item The Achanta et al. version of superpixels \cite{Achanta12} starts from an initial structure in which all superpixels have the same size (homogeneous structure), while the trixels version is created from a heterogeneous perspective. In fact, it is possible the existence of very small trixels (2 or 3 pixels) in TriToM. Due to their small size, those trixels have such a strong centroid which makes them unable to associate with other trixels.
\item Besides, superpixels are allowed to lose their original shape. As they release (or acquire) pixels, their shape may be deformed. For the proposed methodology, trixels are allowed to associate between themselves, but the loss of their individual pixels is not permitted. Our idea states that the segmentation process must preserve the original number of these meaningful regions (trixels).
\end{itemize}

\begin{figure}[tbp]
  \centering
  \includegraphics[width=130mm]{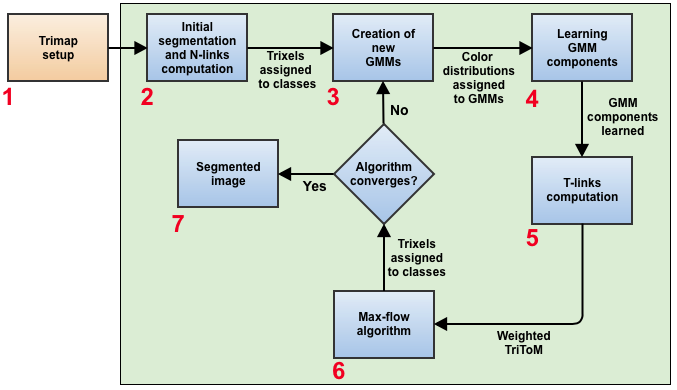}
  \caption{Segmentation flowchart. Block 1 setups the GrabCut trimap, this defines the object of interest (e.g. clothes). The rest of blocks are generic to any segmentation process.}
  \label{fig:flowseg}
\end{figure}

The flowchart of the segmentation process can be observed in Figure \ref{fig:flowseg}. The first box defines what is the object of interest (e.g., skin and clothes in our proposal), while the rest of boxes are generic, no matter what is the object of interest. As can be seen, steps 3 to 6 are repeated 
depending on the algorithm convergence. Moreover, during each iteration, the algorithm is executed until the 
clustering process converges. The following describes the flowchart that allows the 
trixels clustering through the GrabCut algorithm.
\begin{figure}[!htb]
\minipage{0.5\textwidth}
\centering
  \includegraphics[width=40mm]{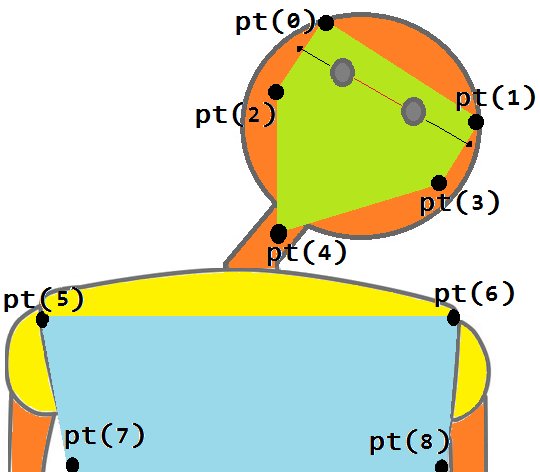}
  \caption{Geometric trimap \cite{Freire14}.}
  \label{fig:geotech}
\endminipage\hfill
\minipage{0.5\textwidth}
\centering
  \includegraphics[width=60mm]{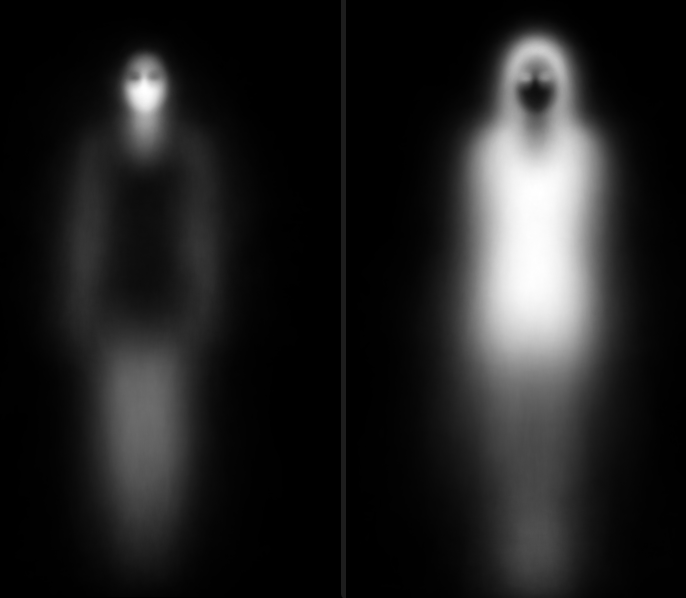}
  \caption{Probabilistic trimap. The left mask shows the skin segmentation trimap. The right mask shows the clothes segmentation trimap. Brighter regions in the image correspond to areas with higher probability.}
  \label{fig:probmap}
\endminipage\hfill
\end{figure}
\begin{enumerate}
\item \textbf{Automatic trimap setup}. The trimap is the input parameter that initializes the probabilistic color models of GrabCut. In our approach, it is defined by the tuple $T=\{T_U, T_F, T_B\}$; where $T_F$ and $T_B$ are the trixels marked as foreground and background respectively and $T_U$ are the remaining trixels. 
As shown in Figure \ref{fig:genderflow}, a face detector provides the necessary cues to define the object of interest. In order to compute this two-step segmentation process, two different trimap definition alternatives are evaluated:
\begin{itemize}
\item \textbf{Geometric trimap}. This trimap definition requires a previous face and eyes detection \cite{Castrillon07-jvci}. Then, both eye locations are used to estimate the different trimap regions based on Equation \ref{eq:geomadapt}. 
\begin{equation}
\forall n \in [0, k], \quad pt(n) = (f_{dist} + w * \vec{v}) + \vec{p_e} * w_n 
\label{eq:geomadapt}
\end{equation}
Where $k$ are the necessary points to estimate each region, $\vec{v}$ represents the vector from one eye to the other, $w$ is the symmetric distribution to achieve equidistant points to both sides of the face, and  $w_n$ is the individual distribution of the weights to move on the y-axis. The vector $\vec{p_e}$ allows the 90 degrees rotation along the image for each point.  The intersection of these points provides a suitable mask for the $T_F$ trimap selection. As can be seen in Figure \ref{fig:geotech}, the green mask is considered for skin segmentation output, while the blue mask is combined with the skin segmentation in order to proceed with the final clothes segmentation step. Therefore, when skin segmentation is considered, green region is $T_F$, white region is $T_B$ and the rest of regions are $T_U$. When clothes segmentation is considered, blue region is $T_F$, white region and the previously segmented skin region are $T_B$ and the rest of regions are $T_U$.

\item \textbf{Probabilistic trimap}. A Bayesian technique that requires a conditional probability map is applied.  Two probability maps were considered, one for skin detection and another for clothes detection (see Figure \ref{fig:probmap}). In our work, each probability map was obtained from the Color-Fashion dataset \citep{LiuSi13-mm}. 
Once again, the facial detector provides both eye locations in the image. Then, each image is 
rotated, re-scaled and translated so that the middle eyes position is placed at a fixed
location $(x_0,y_0)$. According to this location, the different trimap regions are computed based on Equation \ref{eq:prob_pixel}.
\begin{equation}
\label{eq:prob_pixel}
P(L|(x_n,y_n)) = \frac{P(L)P((x_n,y_n)|L)}{P(x_n,y_n)} > \varepsilon  
\end{equation}
Where $L$ is the considered label (foreground, background or unknown), $(x_n,y_n)$ is the normalized pixel position according to $(x_0,y_0)$ and $\varepsilon$ is the decision threshold. See Figure \ref{fig:probmap} for the resulting skin and clothes trimaps.
\end{itemize}

Besides, the segmentation process shown in Figure \ref{fig:flowseg} is applied twice for each person located in the image. First, the skin is segmented and then, the clothes are segmented taking into account the skin region previously segmented. This process helps to avoid the inclusion of skin areas in the clothes segmentation output (e.g. discard skin region when short sleeve or cleavage appears).

\item \textbf{Initial segmentation and N-links computation}. An initial segmentation is computed. In this step, all unknown trixels are tentatively placed in the foreground class.  On the other hand, N-links connect trixels in the 3-neighborhood. These links describe the penalty for placing a segmentation boundary between the neighboring trixels. By applying Equation \ref{eq:nenlace} it is ensured that this penalty is quite high in regions of low gradient and low in regions of high gradient (where real edges are located). Thus, a N-link between trixel $T_i$ and trixel $T_j$ is computed as follows:

\begin{equation}
N(T_i,T_j) = \frac{50}{dist(T_i,T_j)}   \mathrm{e}^{- \beta \| z_{T_i} - z_{T_j} \|^{2}} 
\label{eq:nenlace}
\end{equation}

Where $z_{T_i}$ stands for the mean color of all the pixels contained by trixel $T_i$, $dist(T_i, T_j)$ describes the Euclidean distance between both trixels centroids and $\beta$ considers the distance average of every pair of trixels neighbors ($T_q, T_t$) in the mesh:

\begin{equation}
\beta = \frac{1}{2 \times avg (dist(z_{T_q}, z_{T_t}))}
\label{eq:beta}
\end{equation}

\item \textbf{Creation of new GMMs}. K components of the Gaussian Mixture Models (GMMs) are created for both foreground and background regions, i.e. $2 \times K$. Gaussian components are then initialized from the color distribution in each cluster. For good separation between foreground and background, it is necessary to generate low variance Gaussian components. Secondly, each trixel in the foreground class is assigned to the most likely Gaussian component in the foreground GMM. This is done by evaluating the Gaussian equation with trixels mean color as input. Similarly, each trixel in the background is assigned to the most likely background Gaussian component. Therefore, both regions are divided into K trixel clusters depending on the mean color inside each trixel.

\item \textbf{Learning GMMs components}. The color distribution of each GMMs changes due to the fact that there are trixels which may change from foreground to background during each iteration of the algorithm.
Hence, once the trixels have been clustered, the Gaussian components are updated from the new trixel distributions.

\begin{figure}[tbp]
  \centering
  \includegraphics[width=85mm]{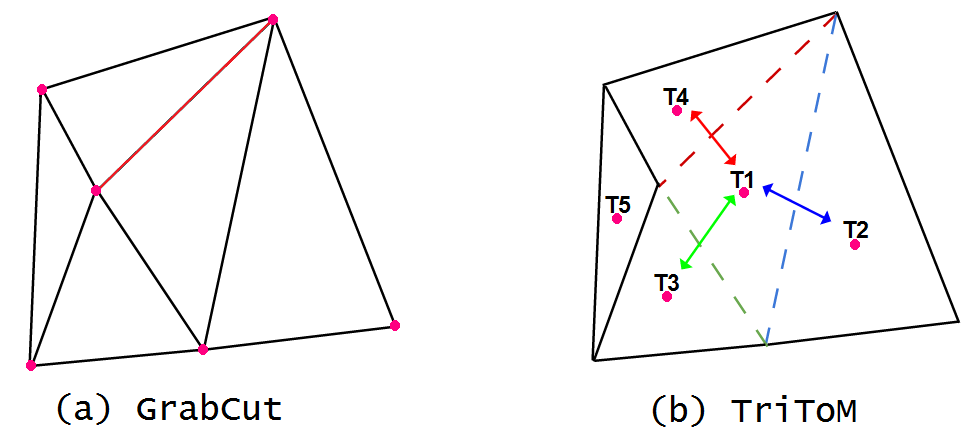}
  \caption{
The fuchsia cores can be seen in both images. These cores provide the necessary information to generate T-links and N-links. In the classical version of GrabCut, each pixel is a core and, therefore, it is also a vertex of the graph which is used to obtain the necessary  information for both types of links. For TriToM, the necessary information  to supply the links is provided by the own trixel. Although edges are used in order to separate the trixels, the links information is computed considering each trixel information, not each vertex information.}
  \label{fig:links}
\end{figure}

\item \textbf{T-links computation}. T-links connect each trixel to both foreground and background classes. Such links describe the probability for a trixel to belong to any of these classes. Furthermore, this probability is computed based on the parameters stored in the previously trained GMM models. 
In addition, in the classical GrabCut version nodes (pixels) are considered vertices and they are connected through edges. Moreover, information contained in both T-links and N-links, arises from the information contained in the vertices themselves (see Figure \ref{fig:links}-a). In contrast, vertices remain only from a physical point of view (non functional) in TriToM. In other words, these vertices maintain the structure that connects the graph edges but the information calculated for T-links and N-links, is not based on the vertices information, but does on the trixel information itself (see Figure \ref{fig:links}-b).

\begin{figure}[t]
\begin{minipage}[b]{1.0\linewidth}
  \centering
  \centerline{\includegraphics[width=8cm]{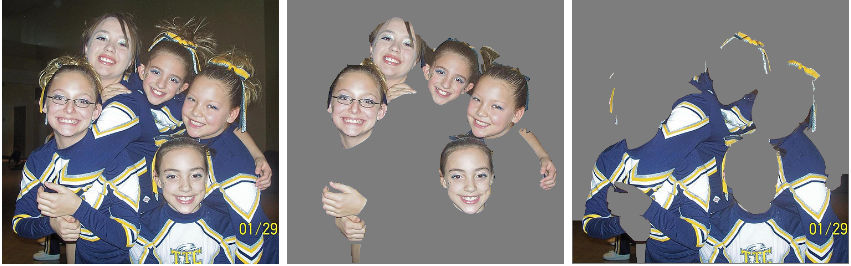}}
\centerline{\includegraphics[width=8cm]{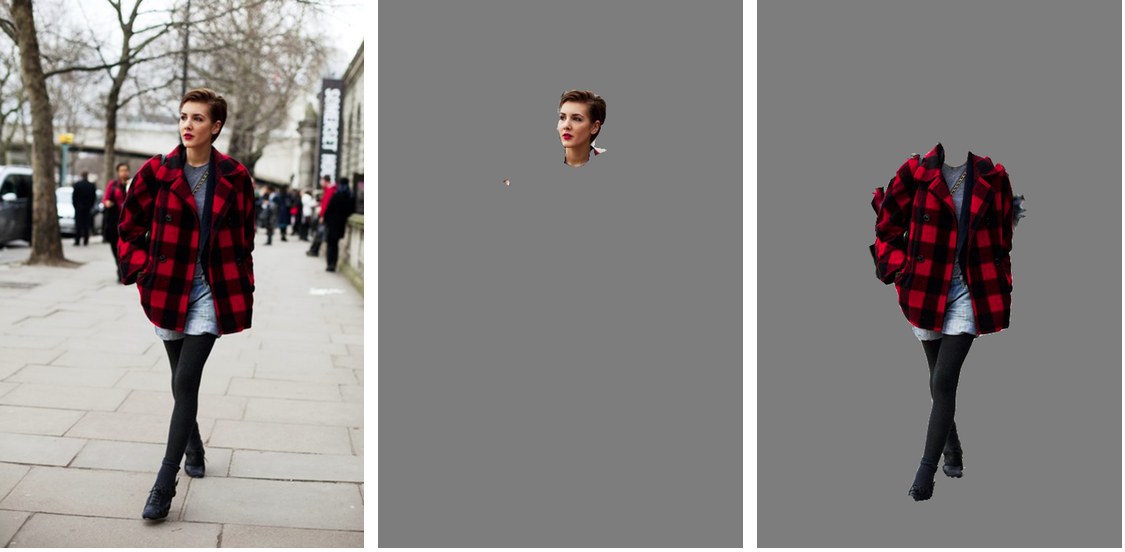}}
\end{minipage}
\caption{Automatic skin (second column) and clothes (third column) extraction from samples of the Pascal dataset \cite{PASCAL_DB} and the ClothesDB \cite{ChenH12-eccv} respectively.}
\label{fig:segexamples}
\end{figure}

\item \textbf{Max-flow algorithm}. This algorithm has been adapted in order to work with trixels instead of pixels, but it follows the same procedure described in \cite{Boykov01}; three stages repeated until each edge between the search trees becomes saturated (an edge is saturated when its N-link value is 0). Trixels which are not in S (foreground cluster) or T (background cluster) are known as free trixels. Trixels in the search trees S and T can be either active or passive. Active trixels represent the outer border in each tree while passive trixels are internal. Besides, active trixels allow trees to grow by acquiring new children (along non-saturated edges) from the set of free trixels. Passive trixels cannot grow as they are completely blocked by other trixels from the same tree. Hence, the three previously mentioned stages are known as \textit{growth} stage, \textit{augmentation} stage and \textit{adoption} stage. In the growth stage, active trixels explore adjacent non-saturated edges in order to find an augmentation path when both search trees encounter. Then, the augmentation stage push the largest flow through the augmentation path, and some edge(s) in the path can become saturated. When this happens, the search trees can be broken into forests. Finally, the adoption stage restores again the original search trees. At the end of the process, the saturated edges provides the segmentation path for each group. Figure \ref{fig:segexamples} shows skin/clothes segmentation examples using GrabCut with trixels.
\end{enumerate}


\section{Feature extraction and classification}
\label{sec:setup}

Local descriptors are often used to obtain pattern information (e.g. LBP can provide a histogram for image data compactness). However, in order to avoid losing information related to the feature location, the image is usually divided into cells \cite{Ahonen06}. We have considered two well known descriptors: LBP \cite{Ojala96-pr} and HOG \cite{Dalal05}. The chosen descriptors provide the possibility to achieve an optimal performance by analyzing both texture (LBP) and edges information (HOG).

\begin{figure}[tbp]
  \centering
  \includegraphics[width=100mm]{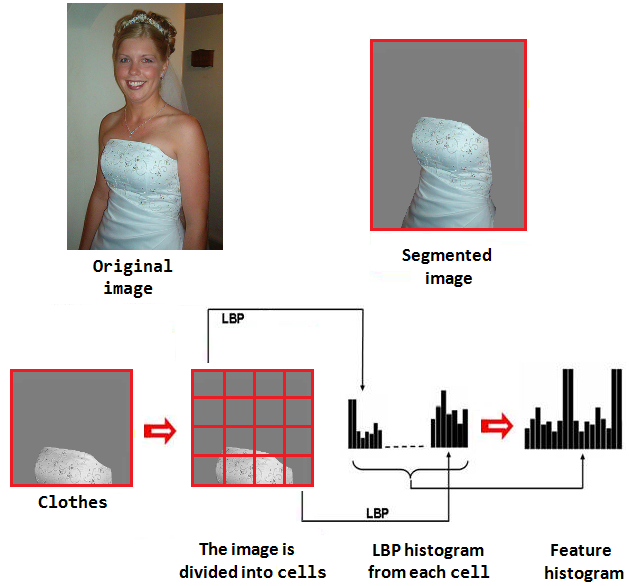}
  \caption{LBP features extraction process.}
  \label{fig:lbpext}
\end{figure}

The LBP is an image descriptor which was initially introduced by Ojala \cite{Ojala96-pr} for texture classification, but its popularity grew because of its invariance under illumination changes and reduced processing cost. Given a grayscale image, a pixel is selected and the LBP operator is applied to its neighbors in order to obtain the LBP pattern. Therefore, a threshold operation is performed in a circular way involving the surrounding pixels of the selected pixel. Then, the result of each of these threshold operations leads to the computation of the binary pattern. 
As can be seen in Figure \ref{fig:lbpext}, first the image is cropped in order to avoid occlusions and to optimize the process. Figure \ref{fig:lbpext} also shows how the resulting image 
is later divided into cells to avoid losing information related to the feature location. Then, the computed LBP patterns are compiled in histograms for each cell. Finally, the descriptor is obtained by the concatenation of each histogram.

\begin{figure}[tbp]
  \centering
  \includegraphics[width=100mm]{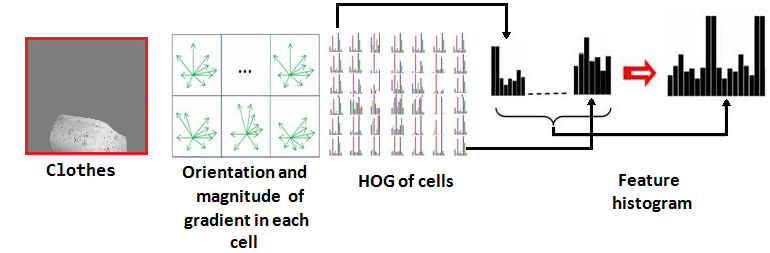}
  \caption{HOG features extraction process.}
  \label{fig:hogext}
\end{figure}

On the other hand, the HOG features are a robust way of describing local object appearances and shapes by their distribution of intensity gradients or edge directions, and have been used successfully as a low level feature in object recognition tasks. In fact, there are works that explicitly exploit the human's shape either holistically \cite{Gavrila07} or in part-based approaches \cite{Lin10}. As can be seen in Figure \ref{fig:hogext}, the image follows the same procedure as in LBP. First, the segmented image is cropped, and the resulting image is later divided into small connected regions known as cells.
Then, a histogram of gradient directions is compiled for each cell. Finally, the descriptor is represented by the concatenation of these histograms. 

Once the feature vector is defined, we have adopted SVM \cite{Vapnik99} with RBF kernel for binary classification (male or female). 


The use of CNN is also considered in our work. The CNN architecture is the one proposed by Levi and Hassner \cite{Levi15} that is composed by three convolutional layers with kernel size of 7x7, 5x5 and 3x3 respectively. Each layer is followed by a ReLU and max pooling stages. Last two layers correspond to full connected ones with size 512 each one.

\section{Experiments}
\label{sec:experimental1}

Our experimental section is divided into two subsections. The first one summarizes the segmentation performance of the proposed trimaps (see section \ref{sec:segmentation}) for both, classical Grabcut and TriToM-based. A well-known database such as the PASCAL database \cite{PASCAL_DB} is considered for this segmentation quality evaluation. 
 
On the second subsection, an experiment is conducted to show our GC proposal performance. For this experiment, the clothes segmentation mask is used as input to be combined with a state-of-the-art gender classifier.

\subsection{Skin and clothes segmentation}
\label{sec:experimental1_1}
The PASCAL database contains images of people with a large variety of poses, backgrounds and scales. Due to the complexity of the database in terms of unrestricted backgrounds and poses, it was estimated as the most appropriate dataset for evaluating the trixels based segmentation. Moreover, only pictures where ENCARA2 \cite{Castrillon07-jvci} detected human faces were considered. 
Thus, the subset contains 1500 images. We adapted our approach to segment skin as well as clothes for this experiment because PASCAL only provides people gound truth masks (clothes and skin).

We completed a quantitative comparison of the classical pixel GrabCut and the TriToM based GrabCut. The tests were made from two points of view: the quality of the people segmentation versus the ground truth annotations, and the processing time. In total four different approaches are evaluated considering the algorithms and the initial trimap setup:
\begin{itemize}
\item PixelGC\_Geo. A classical pixel GrabCut approach based on the computation of a geometric trimap as input.
\item PixelGC\_Prob. A classical pixel GrabCut approach which computes a probabilistic trimap as input.
\item TriToMGC\_Geo. A TriToM GrabCut approach based on the computation of a geometric trimap as input.
\item TriToMGC\_Prob. A TriToM GrabCut approach which computes a probabilistic trimap as input.
\end{itemize}

\begin{figure}
\centering
\includegraphics[width=0.80\linewidth]{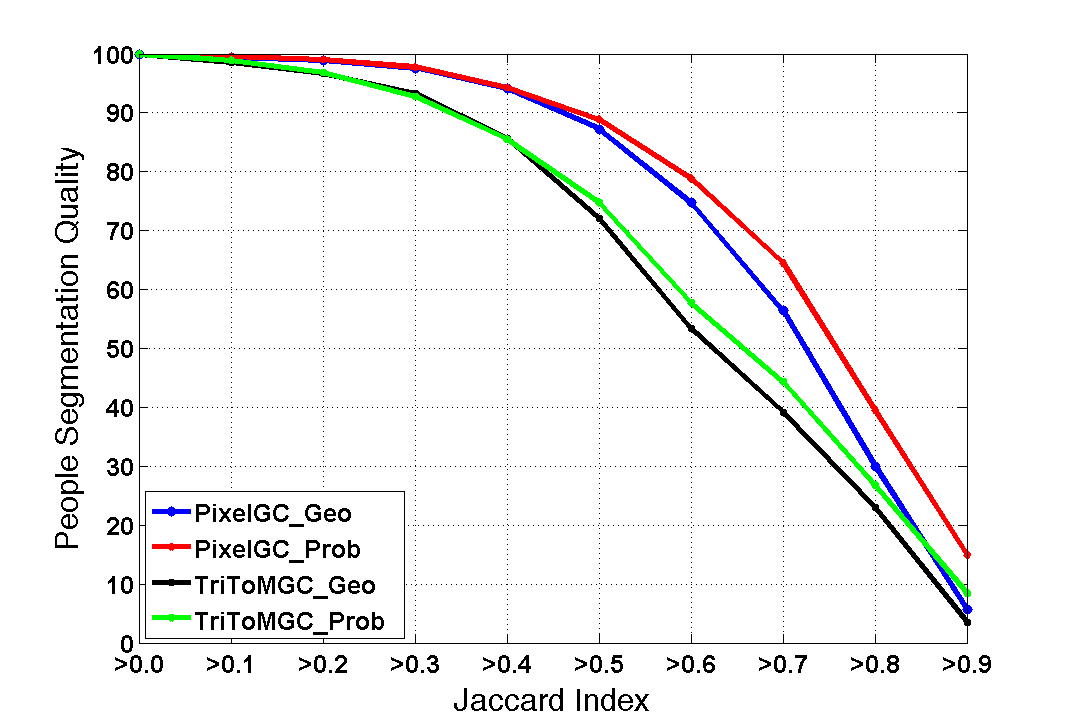}
\vspace{-12pt}
\caption{People segmentation quality vs Jaccard index.}
\label{fig:jaccard-people}
\end{figure}

The Jaccard index (JI) is considered to evaluate the segmentation quality. Let us consider people segmentation as the combination of skin and clothes segmentation. Then, $Seg$ and $GT$ are the result of the people segmentation process and the ground truth segmentation respectively. The JI to estimate the similarity between both segmentation masks can be directly applied by performing Equation \ref{eq:jaccard_eq}.
\begin{equation}
\label{eq:jaccard_eq}
Jaccard\,Index =  \frac{Seg \cap GT}{Seg \cup GT} 
\end{equation}

Figure \ref{fig:jaccard-people} shows the quality of the four different approaches under evaluation. These measures are taken considering the JI of the foreground pixels for each image. As it happens in \cite{Freire14}, the TriToM-based approach does not improve the results of the classical GrabCut approach. As can be also seen in the graph, the probabilistic trimap outperforms the geometric trimap in every case. The improvement depends on the selected JI threshold. For example, choosing a JI threshold of 0.6, an approximately quality improvement of 4\% can be appreciated for both, the PixelGC\_Prob and the TriToMGC\_Prob approaches.

Secondly, the processing cost of the algorithms was studied considering the number of iterations of the segmentation process.

\begin{table}[ht]
\caption{A normalized number of iterations comparison among the approaches.}
\centering
\begin{small}
\centering
 \begin{tabular}{c|c|c|c}
  PixelGC\_Geo & PixelGC\_Prob & TriToMGC\_Geo & TriToMGC\_Prob \\ \hline \hline
  1.0 & 0.9 & 0.2 & 0.2 \\ \hline    
\end{tabular}
\end{small}
\label{tab:speedcomp}
\end{table}

There is a simplification of the process due to the fact that trixels only provide the color distribution inside them \cite{Freire14}. For an example image of $270 \times 349$ pixels, while the classical pixel GrabCut needs $94230$ pixels to work properly, the TriToM-based version only uses 4403 trixels, just 4.6\% of the original inputs. However, this reduction of information has a significant positive side in terms of speed. Table \ref{tab:speedcomp} shows the remarkable improvement over the classical pixel GrabCut. The TriToM GrabCut proposal is roughly a $80\%$ faster than the original GrabCut. In the next subsection, we evaluate whether the segmentation quality decrease achived by the TriToM-based approach is relevant for GC. A similar GC accuracy, would evidence the interest of the TriToM-based approach for faster processing.

\subsection{Gender classification}
\label{sec:genclas}

\begin{figure}[t]
\begin{minipage}[b]{1.0\linewidth}
  \centering
  \centerline{\includegraphics[width=7cm]{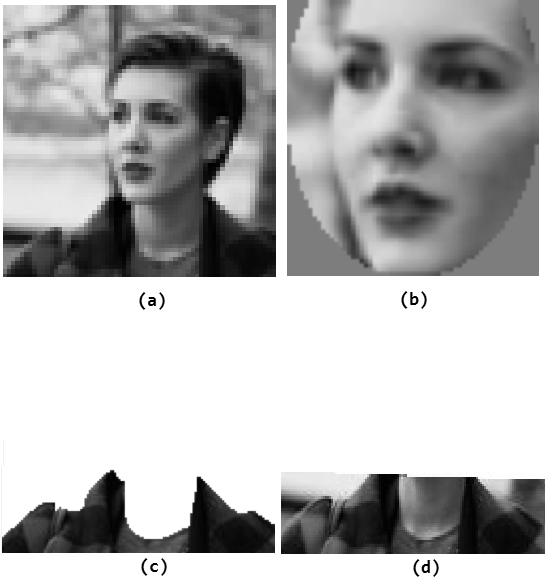}}

\end{minipage}
\caption{Data inputs for feature extraction. As can be appreciable, image (a)  stands for the grayscale head and shoulders image (HS), image (b) stands for the grayscale face image (F), image (c) stands for the grayscale clothes ROI masked (C) and image (d) stands for non masked upper torso ROI (Fixed mask).
The second row of Figure \ref{fig:segexamples} shows the original image.}
\label{fig:classinput}
\end{figure}

An extensive evaluation of the system for the 
GC was done considering the ClothesDB proposed by Chen et al. \cite{ChenH12-eccv}. This database is suitable to our aim because the number of samples per class is higher than other datasets. In this experiment color is no longer required for features extraction (see Figure \ref{fig:classinput}). 

Full body images are hard to extract from surveillance cameras. However, the upper part of the body (head and shoulder) provides regions of the human anatomy that can be feasible extracted by those cameras. In this context, head and shoulders (HS) and clothes (C) can be considered due to the distinct appearance configuration that both male and female can present (e.g. presence of necklace or tie, shoulder contour, neck thickness, etc.). As can be noticed in Figure \ref{fig:classinput}, several cues are considered for the experiment. Both, head and shoulders (HS) and clothes (C) cues have been scaled to $64\times64$ pixels, while the facial image (F) has been scaled to $59\times65$ pixels. 
Then, all cues have been characterized by the local descriptors described in section \ref{sec:setup}: LBP and HOG. For the LBP cell division we conducted an experimental study where the $5\times5$ cell division exhibits the best performance. 
Besides, for the HOG cell division we also conducted an experimental study where the $8\times8$ cell division achieves the best results. These descriptors provide the input for classifiers. Then, a five repetition of a 5-fold cross validation was performed in order to obtain GC rates.

Table 2 summarizes results achieved on the ClothesDB. We include the results by Chen et al. and our previous work \cite{Freire14}. Also different gender classifiers trained with The Images of Groups dataset are included in the comparison. The first one was already evaluated in \cite{Freire14}, and combines facial and local context information \cite{Castrillon13-ciarp}. The second one is a recent update that fuses also CNN outputs \cite{Castrillon16_arxiv}. The CNN is below adopted to be combined with those features automatically extracted from the individual clothes. In this regard, the CNN has been also trained with The Image of Groups dataset for the head and shoulders pattern ($HS_{CNN}$) and face pattern ($F_{CNN}$). As can be seen in the fifth row of Table 2, the CNN rates for GC considering $F_{CNN}$ and $HS_{CNN}$ are 86.5\% and 88.4\% respectively. These rates make sense if we bear in mind that HS classifier requires more cues than the F classifier.

In Freire et al. \cite{Freire14} only the geometric mask was considered.
Due to the segmentation performance obtained in section \ref{sec:experimental1_1}, we have now adopted only the probabilistic mask for both, the classical pixel GrabCut (PixelGC) approach and the TriToM GrabCut (TriToMGC) proposal. 
    

 \begin{table}[ht]
\centering
\begin{small}
\centering
 \begin{tabular}{c|c|c}
   Proposal & Data & Accuracy \\ \hline
    \multirow{3}{*}{\cite{Castrillon13-ciarp}}
   &  $F_{HOG}$ & $78.3\%$ \\
   &  $F_{LBP}$ & $82.5\%$ \\
   &  $HS_{HOG}$ & $75.9\%$ \\ \hline
   \multirow{3}{*}{\cite{Castrillon16_arxiv}}
   &  $HS_{CNN}$+$F_{HOG}$+$F_{LBP}$ & $91.1\%$ \\
   &  $HS_{CNN}$+$F_{HOG}$+$F_{LBP}$+$HS_{HOG}$ & $90\%$ \\ \hline
  \multirow{3}{*}{\cite{ChenH12-eccv}}
   &  $F_{LBP}$ & $71.5\%$ \\
   &  $C_{LBP}$ & $81\%$ \\
   &  $F_{LBP}$+$C_{LBP}$ & $84.9\%$ \\ \hline
     \multirow{3}{*}{\cite{Freire14}}
   &  $C_{LBP+HOG}$ & $67.8\%$ \\ 
   &  $C_{LBP+HOG}$+$HS_{HOG}$ & $79.6\%$ \\ 
   &  $C_{LBP+HOG}$+$HS_{HOG}$+$F_{LBP+HOG}$  & $87.1\%$ \\ \hline \hline
     \multirow{3}{*}{CNN}
   &      $HS_{CNN}$ & $88.4\%$ \\ 
   &      $F_{CNN}$ & $86.5\%$ \\ \hline \hline
        \multirow{5}{*}{Fixed mask}
   &  $C_{LBP+HOG}$ & $72\%$ \\ 
   &  $C_{LBP+HOG}$+$HS_{HOG+CNN}$ & $91.1\%$ \\ 
   &  $C_{LBP+HOG}$+$HS_{HOG+CNN}$+$F_{LBP+HOG}$  & $91\%$ \\
   &  $C_{LBP+HOG}$+$HS_{HOG}$+$F_{LBP+HOG+CNN}$  & $90.7\%$ \\ 
   &  $C_{LBP+HOG}$+$HS_{HOG+CNN}$+$F_{LBP+HOG+CNN}$  & $90.9\%$ \\ \hline
     \multirow{5}{*}{Prob.-PixelGC}
   &  $C_{LBP+HOG}$ & $80.4\%$ \\ 
   &  $C_{LBP+HOG}$+$HS_{HOG+CNN}$ & $93\%$ \\ 
   &  $C_{LBP+HOG}$+$HS_{HOG+CNN}$+$F_{LBP+HOG}$  & $92.9\%$ \\ 
   &  $C_{LBP+HOG}$+$HS_{HOG}$+$F_{LBP+HOG+CNN}$  & $91.1\%$ \\ 
   &  $C_{LBP+HOG}$+$HS_{HOG+CNN}$+$F_{LBP+HOG+CNN}$  & \textbf{94.1\%} \\ \hline
        \multirow{5}{*}{Prob.-TriToMGC}
   &  $C_{LBP+HOG}$ & $78\%$ \\ 
   &  $C_{LBP+HOG}$+$HS_{HOG+CNN}$ & $92.7\%$ \\ 
   &  $C_{LBP+HOG}$+$HS_{HOG+CNN}$+$F_{LBP+HOG}$  & $92.9\%$ \\
   &  $C_{LBP+HOG}$+$HS_{HOG}$+$F_{LBP+HOG+CNN}$  & $90.8\%$ \\  
   &  $C_{LBP+HOG}$+$HS_{HOG+CNN}$+$F_{LBP+HOG+CNN}$  & \textbf{94.2\%} \\ \hline   
   
\end{tabular}
\caption{GC accuracy in ClothesDB. F stands for face, C stands for clothes and HS stands for head and shoulders. The fourth row shows the accuracy of our previous work considering the geometric trimap for the trixels approach. The last two rows show the achieved rates for this paper considering the probabilistic map and both classical or TriToM based GrabCut. }
\end{small}
\label{tab:gender_results}
\end{table}

Similar to Li et al. \cite{Li12}, a fixed mask of the upper body clothing images without faces is considered (see Figure \ref{fig:classinput}-d). This experiment with no prior segmentation has been conducted in order to show the relevance of the clothes segmentation technique widely described in this work. As can be noticed in the sixth row of Table 2, when no segmentation is considered, the results are not better than those achieved in the previous works. For instance, the classifier proposed by Castrill\'on et al. \cite{Castrillon16_arxiv} achieves the same rates without considering any clothes (C) information for classification purposes.  

When segmentation approaches are considered, it can be appreciated on Table 2 that the best results are achieved when several cues are taked into account: C, F, and HS. Again, the classifiers proposed by Castrill\'on et al. \cite{Castrillon16_arxiv, Castrillon13-ciarp} achieves interesting results for F and HS, although no clothes were considered in their works. As happens in Chen et al. \cite{ChenH12-eccv}, the clothing features alone are not good enough to define the gender of a person with a great accuracy due to the fashion culture. 
Furthermore, the use of the 
facial pattern alone reported an interesting accuracy of 91\% when trained with a large dataset, but could not cope with results combining facial and clothing 
information that boosted the GC accuracy up to 94.2\%. 

As can be also seen in Table 2, the integration of both best facial based and clothes based classifiers, reveal an improvement in GC performance. It is noticed that our final combination of patterns and features outperformed previous literature results where clothes and facial-based gender classifiers are combined. When we consider CNN and combine the classifier obtained from each probabilistic approach with state-of-the-art classifier trained with The Image of Groups dataset, the accuracy reports a 7\% better rate than previous works. Table 2 shows this improvement for both, TriToMGC and PixelGC probabilistic approaches.

A final important issue that must be addressed is the similar performance between both PixelGC and TriToMGC proposals. There is no significance difference between both approaches in terms of accuracy but a notable improvement of the TriToMGC proposal over the classical PixelGC approach when the processing time is considered.

\section{Conclusions}
\label{sec:conclusions}

In this work, we have improved GC making use of a novel segmentation technique based on the integration of two complementary techniques; trixels and GrabCut. TriToM simplifies the image data into perceptually meaningful atomic regions known as trixels. Then, our TriToMGC approach segments the trixel based image instead of the classical pixels configuration. 

The optimized automatic clothes segmentation 
approach close to real-time adapts GrabCut to use TriToM as input. In this sense, a quantitative comparison within PASCAL, reported a 6\% worse accuracy, but requiring just 20\% of the processing time. Our trixel based segmentation demonstrates faster processing on unconstrained images, by using the TriToM-based GrabCut algorithm. 

Later, this focus has been adopted for clothes extraction, with the final goal at being integrated in a GC system. State-of-the-art facial based GC systems have shown a remarkable performance combining local descriptors and CNN. Our study have evidenced that the fusion with features extracted from the clothes achieves a significant improvement in GC accuracy. The resulting GC accuracy is similar for both pixel or trixel based segmentation, suggesting that trixels based clothes segmentation, being faster, provides enough segmentation quality for GC.

\section*{References}

\bibliography{mybibfile}

\begin{thebibliography}{10}
\expandafter\ifx\csname url\endcsname\relax
  \def\url#1{\texttt{#1}}\fi
\expandafter\ifx\csname urlprefix\endcsname\relax\def\urlprefix{URL }\fi
\expandafter\ifx\csname href\endcsname\relax
  \def\href#1#2{#2} \def\path#1{#1}\fi

\bibitem{Dantcheva16}
A.~Dantcheva, P.~Elia, A.~Ross, What else does your biometric data reveal? a
  survey on soft biometrics, IEEE Transactions on Information Forensics and
  Security 11~(3) (2016) 441--467.

\bibitem{Bhardwaj15-icb}
R.~Bhardwaj, G.~Goswami, R.~Singh, M.~Vatsa, Harnessing social context for
  improved face recognition, in: International Conference on Biometrics (ICB),
  2015, pp. 121--126.

\bibitem{Li12}
B.~Li, X.~Lian, B.~Lu, Gender classification by combining clothing, hair and
  facial component classifiers, Neurocomputing 1~(76) (2012) 18--27.

\bibitem{Choon12}
C.~B. Ng, Y.~H. Tay, B.-M. Goi, Vision-based human gender recognition: A
  survey, CoRR abs/1204.1611 (2012) 26---33.

\bibitem{Rother04}
C.~Rother, V.~Kolmogorov, A.~Blake, Grabcut - interactive foreground extraction
  using iterated graph cuts, Proceedings of ACM {SIGGRAPH} (2004) 309--314.

\bibitem{Haasch05}
A.~Haasch, N.~Hofemann, J.~Fritsch, G.~Sagerer, A multi-modal object attention
  system for a mobile robot, Proceedings of the of the International Conference
  on Intelligent Robots and Systems (2005) 1499--1504.

\bibitem{Vaiapury10}
K.~Vaiapury, A.~Aksay, E.~Izquierdo, Grabcutd: Improved grabcut using depth
  information, Proceedings of the 2010 ACM Workshop on Surreal Media and
  Virtual Cloning (2010) 57--62.

\bibitem{Freire14}
D.~Freire-Obreg\'on, M.~Castrill\'on-Santana, J.~Lorenzo-Navarro,
  E.~Ram\'on-Balmaseda, Automatic clothes segmentation for soft biometrics,
  Proceedings of the IEEE International Conference on Image Processing (ICIP)
  (2014) 4972-- 4976.

\bibitem{Rai14}
P.~Rai, P.~Khanna, A gender classification system robust to occlusion using
  gabor features based (2d) pca, Journal of Visual Communication and Image
  Representation 25~(5) (2014) 1118--1129.

\bibitem{Wu11}
X.~Wu, X.~Lian, Multi-view gender classification using symmetry of facial
  images, Neural Computing and Applications (2011) 1--9.

\bibitem{Shan12}
C.~Shan, Learning local binary patterns for gender classification on real world
  face images, Pattern Recognition Letters 33~(4) (2012) 431---437.

\bibitem{Castrillon16_arxiv}
M.~Castrill{\'o}n-Santana, J.~Lorenzo-Navarro, E.~Ram{\'o}n-Balmaseda,
  Descriptors and regions of interest fusion for in- and cross-database gender
  classification in the wild, Image and Vision Computing (in press).

\bibitem{Cao08}
L.~Cao, Y.~Fu, M.~Dikmen, T.~Huang, Gender recognition from body, Proceedings
  of the 16th ACM International Conference on Multimedia (2008) 725--728.

\bibitem{Collins09}
M.~Collins, J.~Zhang, P.~Miller, Full body image feature representations for
  gender profiling, Proceedings of the IEEE 12th International Conference on
  Computer Vision Workshops (2009) 1235--1242.

\bibitem{Bourdev11}
L.~Bourdev, S.~Maji, J.~Malik, Describing people: A poselet-based approach to
  attribute classification., IEEE International Conference on Computer Vision
  (ICCV) (2011) 1543--1550.

\bibitem{Ren03}
X.~Ren, J.~Malik, Learning a classification model for segmentation, Proceeding
  Proceedings of the Ninth IEEE International Conference on Computer Vision
  1~(1) (2003) 1--8.

\bibitem{Achanta12}
R.~Achanta, A.~Shaji, K.~Smith, A.~Lucchi, P.~Fua, S.~Susstrunk, {SLIC}
  superpixels compared to state-of-the-art superpixel methods, IEEE
  Transactions on Pattern Analysis and Machine Intelligence 34~(11) (2012)
  2274--2282.

\bibitem{Levinshtein13}
A.~Levinshtein, C.~Sminchisescu, S.~Dickinson, Multiscale symmetric part
  detection and grouping, International Journal of Computer Vision 104~(2)
  (2013) 117--134.

\bibitem{Fulkerson09}
B.~Fulkerson, A.~Vedaldi, S.~Soatto, Class segmentation and object localization
  with superpixel neighborhoods, Proceedings of the IEEE 12th International
  Conference on Computer Vision 1~(1) (2009) 670--677.

\bibitem{Rosenfeld68}
A.~Rosenfeld, J.~Pfaltz, Distance functions on digital pictures, Pattern
  Recognition 1~(1) (1968) 33--61.

\bibitem{Anton12}
L.~Ant\'on-Canal\'is, M.~Hern\'andez-Tejera, E.~S\'anchez-Nielsen, Distance
  maps from unthresholded magnitudes, Pattern Recognition 45~(9) (2012)
  3125--3130.

\bibitem{Castrillon07-jvci}
M.~Castrill\'on-Santana, O.~D\'eniz-Su\'arez, M.~Hern\'andez-Tejera,
  C.~Guerra-Artal, {ENCARA2}: Real-time detection of multiple faces at
  different resolutions in video streams, Journal of Visual Communication and
  Image Representation (2007) 130--140.

\bibitem{LiuSi13-mm}
S.~Liu, J.~Feng, C.~Domokos, H.~Xu, J.~Huang, Z.~Hu, S.~Yan, Fashion parsing
  with weak color-category labels, IEEE Transactions on Multimedia 16~(1)
  (2014) 253--265.

\bibitem{PASCAL_DB}
C.~Williams, J.~Winn, M.~Everingham, L.~Gool, A.~Zisserman, The pascal visual
  object classes (voc) challenge, International Journal of Computer Vision 88
  (2009) 303--338.

\bibitem{ChenH12-eccv}
H.~Chen, A.~Gallagher, B.~Girod, Describing clothing by semantic attributes,
  Proceedings of the European Conference on Computer Vision (ECCV) (2012)
  609---623.

\bibitem{Boykov01}
Y.~Boykov, M.~Joly, Interactive graph cuts for optimal boundary and region
  segmentation of objects in n-d images, IEEE International Conference on
  Computer Vision (ICCV) (2001) 105--112.

\bibitem{Ahonen06}
T.~Ahonen, A.~Hadid, M.~Pietik{\"a}inen, Face description with local binary
  patterns: Application to face recognition, IEEE Transactions on Pattern
  Analysis and Machine Intelligence 28~(12) (2006) 2037--2041.

\bibitem{Ojala96-pr}
T.~Ojala, M.~Pietik{\"a}inen, D.~Harwood, A comparative study of texture
  measures with classification based on featured distributions, Pattern
  Recognition 29 (1996) 51--59.

\bibitem{Dalal05}
N.~Dalal, B.~Triggs, Histograms of oriented gradients for human detection,
  Proceedings of the IEEE Computer Society Conference on Computer Vision and
  Pattern Recognition 1 (2005) 886--893.

\bibitem{Gavrila07}
D.~Gavrila, A bayesian, exemplar-based approach to hierarchical shape matching,
  IEEE Transactions on Pattern Analysis and Machine Intelligence 29~(8) (2007)
  1408--1421.

\bibitem{Lin10}
Z.~Lin, L.~Davis, Shape-based human detection and segmentation via hierarchical
  part template matching, IEEE Transactions on Pattern Analysis and Machine
  Intelligence 32~(4) (2010) 604--618.

\bibitem{Vapnik99}
V.~Vapnik, The Nature of Statistical Learning Theory, Springer-Verlag New York,
  Inc., 1995.

\bibitem{Levi15}
G.~Levi, T.~Hassner, Age and gender classification using convolutional neural
  networks, IEEE Computer Society, 2015.

\bibitem{Castrillon13-ciarp}
M.~Castrill\'on, J.~Lorenzo, E.~Ram\'on, Improving gender classification
  accuracy in the wild, Proceedings of the 18th Iberoamerican Congress on
  Pattern Recognition (CIARP) (2013) 270--277.

\end{thebibliography}

\end{document}